\documentclass[10pt,twocolumn,letterpaper]{article}

\usepackage[pagenumbers]{cvpr}

\usepackage{graphicx}
\usepackage{amsmath}
\usepackage{amssymb}
\usepackage{booktabs}

\makeatletter
\@namedef{ver@everyshi.sty}{}
\makeatother

\usepackage[pagebackref,breaklinks,colorlinks]{hyperref}

\usepackage[capitalize]{cleveref}
\crefname{section}{Sec.}{Secs.}
\Crefname{section}{Section}{Sections}
\Crefname{table}{Table}{Tables}
\crefname{table}{Tab.}{Tabs.}

\DeclareMathOperator{\R}{\mathbb{R}}
\DeclareMathOperator{\N}{\mathcal{N}}
\DeclareMathOperator{\D}{\mathcal{D}}
\DeclareMathOperator{\ypred}{\mathnormal{\hat{y}}}
\DeclareMathOperator{\med}{\mathrm{t}}
\DeclareMathOperator{\mae}{\mathrm{s}}
\DeclareMathOperator{\likeli}{\mathcal{L}}
\DeclareMathOperator{\E}{\mathbb{E}}

\let\b\relax

\DeclareMathOperator{\x}{\mathnormal{x}}
\DeclareMathOperator{\y}{\mathnormal{y}}
\DeclareMathOperator{\W}{\mathnormal{W}}
\DeclareMathOperator{\DW}{\mathnormal{\Delta W}}
\DeclareMathOperator{\b}{\mathnormal{b}}
\DeclareMathOperator{\A}{\mathnormal{A}}
\DeclareMathOperator{\B}{\mathnormal{B}}
\DeclareMathOperator{\cin}{\mathnormal{c}_{\mathrm{in}}}
\DeclareMathOperator{\cout}{\mathnormal{c}_{\mathrm{out}}}
\DeclareMathOperator{\din}{\mathnormal{d}_{\mathrm{in}}}
\DeclareMathOperator{\dout}{\mathnormal{d}_{\mathrm{out}}}

\newcommand{\eq}[2]{\begin{gather}\begin{aligned} \label{eq:#1} #2 \end{aligned}\end{gather}}

\usepackage[dvipsnames]{xcolor}

\hypersetup{%
  linkcolor   = PineGreen,
  urlcolor    = PineGreen,
  citecolor   = PineGreen
}

\def\cvprPaperID{28} 
\def\confName{UNCV-W}
\def\confYear{2024}

\begin{document}

\title{Parameter-efficient Bayesian Neural Networks \\ for Uncertainty-aware Depth Estimation}

\author{
Richard D. Paul$^{1,2,*}$ $\quad$ Alessio Quercia$^{1,3,*}$ $\quad$ Vincent Fortuin$^{2,4}$  $\quad$ Katharina N\"oh$^1$ $\quad$ Hanno Scharr$^1$ \\[1mm]
$^1$ Forschungszentrum J\"ulich, J\"ulich, Germany \\
$^2$ Helmholtz AI, Munich, Germany \\
$^3$ RWTH Aachen University, Aachen, Germany \\
$^4$ Technical University of Munich, Munich, Germany \\
{\tt\small \{r.paul,a.quercia,k.noeh,h.scharr\}@fz-juelich.de} \\[-1mm]
{\tt\small vincent.fortuin@tum.de}
}

\maketitle

\begin{abstract}
    State-of-the-art computer vision tasks, like monocular depth estimation (MDE), rely heavily on large, 
    modern Transformer-based architectures.
    However, their application in safety-critical domains demands reliable predictive performance
    and uncertainty quantification. 
    While Bayesian neural networks provide a conceptually simple approach to serve those requirements, 
    they suffer from the high dimensionality of the parameter space.
    Parameter-efficient fine-tuning (PEFT) methods, in particular low-rank adaptations (LoRA),
    have emerged as a popular strategy for adapting 
    large-scale models to down-stream tasks by performing parameter inference on lower-dimensional
    subspaces. 
    In this work, we investigate the suitability of PEFT methods for subspace Bayesian inference
    in large-scale Transformer-based vision models. 
    We show that, indeed, combining BitFit, DiffFit, LoRA, and CoLoRA, 
    a novel LoRA-inspired PEFT method, with Bayesian inference 
    enables more robust and reliable predictive performance in MDE.
\end{abstract}

\section{Introduction}
\label{sec:intro}

Recent years have seen the emergence of large-scale self-supervised foundation models
in various domains, especially in computer vision 
\cite{oquab_dinov2_2023,kirillov_segment_2023,yang_depth_2024} 
and natural language processing 
\cite{brown_language_2020, radford_learning_2021, touvron_llama_2023}.
By leveraging large amounts of unlabeled data, these models exhibit remarkable performance, even
under distribution shifts \cite{radford_learning_2021, mayilvahanan_does_2024}. 
Nevertheless, their application in safety-critical environments, such as autonomous driving or healthcare, demands for uncertainty estimation in order to detect distribution shifts
and, thus, enhance the reliability of the model's predictions.
Bayesian deep learning provides a conceptually attractive approach to quantify epistemic uncertainties,
however, due to the large number of parameters, many existing methods become prohibitively expensive \cite{papamarkou_position_2024}.

Although the large growth in model size over the recent years has boosted predictive accuracy
in many domains, it also introduced serious issues regarding the accessibility of such
methods, as their training typically requires large computing infrastructure.
To this end, PEFT methods like LoRA \cite{hu_lora_2021}, BitFit \cite{zaken_bitfit_2022}, 
and DiffFit \cite{xie_difffit_2023} have been proposed, which construct parameter subspaces much smaller than the
original parameter spaces, yet allowing for competitive performance on downstream tasks while 
requiring much less computational power.
Thus, the question arises whether the PEFT subspaces are also suitable for performing less expensive,
yet effective, uncertainty estimation.

The applicability of PEFT subspaces for performing Bayesian inference has so far only started to be
investigated \cite{yang_bayesian_2024,onal_gaussian_2024}.
In particular, Yang et al. \cite{yang_bayesian_2024} applied Laplace approximations to the fine-tuned LoRA adapters
to achieve improved calibration in large-language models.
Onal et al. \cite{onal_gaussian_2024} investigate using Stochastic Weight Averaging 
Gaussians (SWAG) \cite{maddox_simple_2019} with LoRA and find it being competitive to
Laplace approximations.

\subsection{Contribution}
\label{sec:contribution}

In this work, we investigate the suitability of different PEFT subspaces for post-hoc Bayesian inference 
on the state-of-the-art vision foundation model \emph{Depth Anything} \cite{yang_depth_2024} for MDE
using SWAG and checkpoint ensembles \cite{chen_checkpoint_2017}.
In particular, we incorporate the BitFit \cite{zaken_bitfit_2022} and DiffFit \cite{xie_difffit_2023} PEFT methods into our analysis,
which have not yet been investigated for subspace Bayesian inference.

Moreover, since the architecture at hand uses a convolution-based prediction head on top of a vision transformer
backbone \cite{yang_depth_2024}, we propose the construction of a parameter-efficient subspace for convolutional kernels, 
which we call CoLoRA (\Cref{sec:colora}). 
CoLoRA mimics LoRA by applying a low-rank decomposed perturbation based on the Tucker decomposition \cite{hitchcock_expression_1927, tucker_mathematical_1966}.

We find that the PEFT methods under consideration allow for parameter-efficient
Bayesian inference in large-scale vision models for MDE.

\section{Background}
\label{sec:background}

In fine-tuning, we consider a typical supervised learning setting with data $\D = \{(x_i, y_i)\}$.
For MDE, the inputs $x_i \in \R^{3 \times h \times w}$ are RBG images and the outputs $y_i \in \R_+^{h \times w}$
are the depth maps.
Within this work, we consider the depth maps to be in disparity space, instead of metric space.
The disparity of a pixel is obtained as the inverse metric depth of said pixel.
Yang et al. \cite{yang_depth_2024} perform fine-tuning of a pre-trained neural network $f : (x, \theta) \mapsto y$ with parameters $\theta$
by minimizing the \emph{affine-invariant mean absolute error} \cite{ranftl_towards_2022}
\eq{aimae}{
    \ell_{\D}(\theta) = \frac{1}{N} \sum_{i=1}^N \left| \frac{\ypred_i - \med(\ypred_i)}{\mae(\ypred_i)} - \frac{y_i - \med(y_i)}{\mae(y_i)} \right|,
}
where $\ypred_i = f(x_i, \theta)$ is the network prediction, 
$\med(y_i)$ is the spatial median,
and 
$
\mae(y_i) = \frac{1}{hw} \sum_{j=1}^{h\times w} | y_j - \med(y) |.
$ 

\subsection{Bayesian Deep Learning}
\label{sec:bayes}

In Bayesian deep learning, prediction is performed with respect to the parameter posterior distribution
\eq{param-posterior}{
    \pi(\theta | \D) \propto \likeli(\D | \theta) \, p(\theta),
}
where $\likeli(\D | \theta) \propto \exp{-\ell_{\D}(\theta)}$ is the likelihood and $p(\theta)$ the prior.
For prediction on a new data point $x^*$ with true label $y^*$, 
we consider the posterior predictive distribution
\eq{pred-posterior}{
    \pi^*(y^* | x^*, \D) = \E_{\theta \sim \pi}[\likeli(y^* | \theta, x^*)].
}

As the posterior is usually intractable, samples from it need to be approximated to perform Monte-Carlo estimation 
\eq{monte-carlo}{
    \E_{\theta \sim \pi}[h(\theta)] \approx \frac{1}{n} \sum_{i=1}^n h(\theta_i), \quad  \theta_i \sim q(\theta) \approx \pi(\theta | \D)
}
for $h(\cdot)$ being, e.g., the predictive mean or variance.

\subsection{Stochastic Weight Averaging Gaussians \& Checkpoint Ensembles}
\label{sec:swag}

SWAG \cite{maddox_simple_2019} was introduced as a simple baseline for
Bayesian inference by computing Gaussian approximate posteriors from the checkpoints of a standard SGD
training run. 
Given a set of checkpoints reshaped as flattened vectors $\theta_1, \ldots, \theta_n$ with $\theta_i \in \R^d$,
one may choose between a diagonal approximation, in which case the approximate posterior is
$\N(\mu_{\theta}, \sigma_{\theta}^2I)$, where $\mu_{\theta} = \frac{1}{n} \sum_{i=1}^n \theta_i$ and
$\sigma_{\theta}^2 = (\frac{1}{n}\sum_{i=1}^n \theta_i^2) - \mu_{\theta}^2 $,
or a low-rank plus diagonal approximation, in which case the approximate posterior becomes $\N(\mu_{\theta}, \Sigma_{\theta})$,
where $\Sigma_{\theta} = \frac{1}{2}\cdot (\sigma_{\theta}^2I + \Sigma_{\mathrm{lr}})$ 
and $\Sigma_{\mathrm{lr}} = \frac{1}{n-1} \sum_{i=1}^n (\theta_i - \mu_{\theta})(\theta_i - \mu_{\theta})^\top$,
which is low-rank if $n < d$.

Alternatively, instead of computing the moments of the empirical distribution of 
parameter values and then sampling from the corresponding normal distribution to compute
the Monte-Carlo estimate from \Cref{eq:monte-carlo}, one may directly treat the checkpoints
as samples from the posterior distribution.
This method is known as checkpoint ensemble \cite{chen_checkpoint_2017}.

\subsection{LoRA}
\label{sec:lora}

LoRA \cite{hu_lora_2021} is a PEFT strategy, 
which adds low-rank perturbations $\DW$ to large weight matrices $\W$ of linear layers
in modern neural network architectures.
That is, for a linear weight matrix $\W \in \R^{\din \times \dout}$ and bias vector
$\b \in \R^{\dout}$, we compute
\eq{linear-layer}{
    \y = \x \W + \x \DW + \b = \x \W + \x \A\!\B + \b,
}
where $\A \in \R^{\din \times r}, \B \in \R^{r \times \dout}$ 
are factors decomposing $\DW$.
By choosing sufficiently small rank $r$ for $\A$ and $\B$,
we then obtain a low-rank approximation $\DW$.
Fine-tuning is then performed by only optimizing the factors
$\A$ and $\B$.

\subsection{BitFit \& DiffFit}
\label{sec:bitfit}

BitFit \cite{zaken_bitfit_2022} is an alternative PEFT strategy, which only unfreezes the biases of a 
pre-trained model for fine-tuning.
It was shown to be competitive or sometimes even better than performing full fine-tuning on 
language models \cite{zaken_bitfit_2022}.

DiffFit \cite{xie_difffit_2023} extends BitFit by adding additional scalar factors to the attention
and feed-forward blocks of a transformer, as well as unfreezing the layer norms.
In their paper, the authors demonstrate improved performance over BitFit, LoRA, and full fine-tuning 
for diffusion transformers.

\section{Method}
\label{sec:method}

\subsection{CoLoRA}
\label{sec:colora}

The convolution operation in CNNs consists of applying a convolutional kernel $\W$,
which is a tensor of size $\cout \times \cin \times k_1 \times \cdots \times k_d$,
to an input tensor $x$ of size $\cin \times h_1 \times \cdots \times h_d$, before
applying a further additive bias $\b \in \R^{\cout}$.
That is, for every output channel
$i$, we obtain
$
    \y_i = \b_i + \sum_{j=1}^{\cin} h(\x, \rho)_j \ast_{\delta} \W_{ij},
$
where $\ast_{\delta}$ is the cross-correlation operation with stride $\delta$ and
$h(x, \rho)$ is the input signal after applying padding $\rho$ to it.
Leveraging the distributivity of cross-correlations, 
we mimic LoRA for convolutions by considering an additive perturbation
$\DW$ on a given weight matrix $\W$ as
\eq{colora}{
    \y_i = \b_i &+ \sum_{j=1}^{\cin} h(\x, \rho)_j \ast_{\delta} (\W_{ij} + \DW_{ij}) \\
         = \b_i &+ \sum_{j=1}^{\cin} h(\x, \rho)_j \ast_{\delta} \W_{ij} \\
         &+ \sum_{j=1}^{\cin}  h(\x, \rho)_j \ast_{\delta} \DW_{ij}.
}
We then obtain a low-rank
decomposition with core $C$ and factors $U^{(1)}, U^{(2)}$
by applying the Tucker-2 decomposition (cref{sec:tucker}) \cite{hitchcock_expression_1927, tucker_mathematical_1966} on the channel dimensions of size $\cin$ and $\cout$,
as in practice the kernel dimensions $k_1, \cdots, k_d$ are typically much smaller.
As in LoRA, for fine-tuning, we only consider the decomposition 
$C, U^{(1)}, U^{(2)}$.
Moreover, we initialize the low-rank factors such that $\DW$ is zero initially by
setting $U^{(1)}$ to be all zeros and initialize $C$ and $U^{(2)}$ randomly from
a Gaussian distribution with zero mean and the variance computed as in the Glorot initialization \cite{glorot_understanding_2010}.

As presented by Kim et al. \cite{kim_compression_2016}~and Astrid et al. \cite{astrid_cp-decomposition_2017}, 
convolutions with Tucker-2-decomposed kernels can themselves be decomposed into
a sequence of convolutions, where the convolutional kernels are given from the
decomposition.
For $\DW = C \times_{\!1} U^{(1)} \times_{\!2} U^{(2)}$,
we decompose
\eq{conv-decomposed}{
    h(x, \rho) \ast_{\delta} \DW = (h(x \ast_{1} \tilde{U}^{(2)}, \rho) \ast_{\delta} C ) \ast_{1} \tilde{U}^{(1)},
}
where $\tilde{U}^{(1)}, \tilde{U}^{(2)}$ are the unsqueezed factors of
$U^{(1)}, U^{(2)}$ with size $\cout \times r \times 1 \cdots \times 1$ and 
$\cin \times r \times 1 \cdots \times 1$, respectively.
This way, the computation of $h(x, \rho) \ast_{\delta} \DW$ does not require the allocation of $\DW$
during training.

\subsection{Inference}
\label{sec:inference}

We perform Bayesian inference using SWAG and checkpoint ensembles on the parameter-efficient 
subspaces constructed by either BitFit, DiffFit, LoRA, or CoLoRA.
The latter two methods provide an additional rank parameter, for which we consider different rank parameters between 1 and 64.
We start from fine-tuned checkpoints in order to demonstrate the applicability of SWAG and checkpoint ensembles
for post-hoc \emph{Bayesification} of an existing pipeline without any needs for sacrifices in accuracy.
From a Bayesian perspective, such an already fine-tuned checkpoint can be considered as the MAP or MLE estimate,
depending on whether regularization was used during training.
Besides, starting the inference from such a high-density region is recommended for many sampling algorithms,
often called a warm start \cite{gelman_bayesian_2020}.

\section{Experiments}
\label{sec:experiments}

We perform experiments based on the pipeline provided by Yang et al. \cite{yang_depth_2024},
from which we extract the DINOv2 feature encoder \cite{oquab_dinov2_2023} and DPT decoder \cite{ranftl_vision_2021}.
As mentioned in the previous \Cref{sec:inference}, instead of performing real fine-tuning,
we rather act as if continuing fine-tuning, while recording checkpoints, in order
to construct checkpoint ensembles and estimate the first and second moments required for SWAG.
Following Yang et al. \cite{yang_depth_2024}, we fine-tune the model on the popular NYU \cite{silberman_indoor_2012}
and KITTI \cite{geiger_vision_2013} data sets on the very same data splits and using the loss from \Cref{eq:aimae}.

We test the four PEFT methods LoRA, CoLoRA, BitFit, and DiffFit
using four different posterior approximation methods: Deep Ensembles (DeepEns), 
Checkpoint Ensembles (CkptEns),
diagonal SWAG (SWAG-D), and low-rank plus diagonal SWAG (SWAG-LR).
Details on sampling and ranks under consideration are given in \Cref{sec:details}.
We further consider full fine-tuning on all parameters.
As as baseline, we consider the performance obtained from the provided, fine-tuned checkpoints from
Yang et al. \cite{yang_depth_2024}.
All of the following evaluations were performed on the same test splits of the 
NYU and KITTI data sets consisting of 130 randomly drawn images from each data set.

\subsection{Predictive Performance}
\label{sec:performance}

\begin{figure}
    \centering
    \includegraphics[width=\columnwidth]{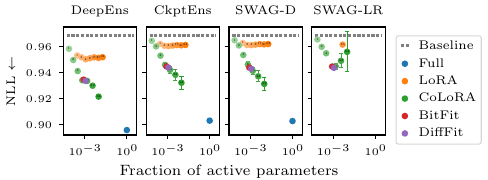}
    \caption{
        Negative log-likelihood for all combinations of inference
        and PEFT methods under consideration, evaluated on the NYU data set. 
        Except SWAG-LR, all methods achieve improved NLL over the deterministic baseline.
        Error bars indicate 95\% intervals across 5 replicate runs.
        Numbers in the dots indicate the rank parameter used.
    }
    \label{fig:nyu-error}
\end{figure}

We measure predictive performance by evaluating the negative log-likelihood (NLL)
on an unseen test split for NYU and KITTI.
We report the results in \Cref{fig:nyu-error} and \Cref{fig:kitti-error}.
We observe improvements when using Bayesian inference on either
the full parameter space or just a PEFT subspace.
Performing inference on the full parameter space achieves the best performance
across all inference methods, except low-rank plus diagonal SWAG (SWAG-LR),
where the evaluation failed due to numerical issues.
For inference on the PEFT subspaces, we observe the most improvements
using CoLoRA with a rank of at least 16, after which it begins to outperform
BitFit and DiffFit, when using DeepEns, CkptEns, or SWAG-D.
Most remarkably, CoLoRA seems to interpolate between the deterministic baseline 
and the full parameter space, if the rank parameter is increased.
For LoRA, we observe improvement over the baseline already for rank 1.
However, quite surprisingly, increasing the rank seems to yield only a small further improvement.

In terms of inference methods, we observe that DeepEns, although consisting only of 5 
different parameter samples, yields better performance than all other methods, where prediction
is performed using 100 parameter samples.
Moreover, we observe little difference between SWAG-D and CkptEns.
For SWAG-LR, we had numerical issues on the NYU data set, resulting
in degraded performance for CoLoRA at ranks above 16, as well as missing evaluation
for LoRA (except rank 16).
For KITTI, we observe similar results for CkptEns and SWAG-D (c.f.,~\Cref{fig:kitti-error}).

\subsection{Calibration}
\label{sec:selective}

\begin{figure}
    \centering
    \includegraphics[width=\columnwidth]{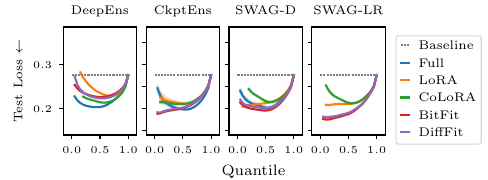}
    \caption{
        Test loss per quantile of most certain predictions evaluated on
        the NYU data set.
        Except DeepEns, all methods achieve improved test loss on more certain
        pixels, suggesting good calibration.
        Uncertainty was estimated using pixelwise standard deviation.
        For LoRA and CoLoRA, only the results for the rank with lowest test loss on the
        5\% quantile are depicted.
        The prediction using the publicly available checkpoint was used as
        a baseline.
        Shaded areas indicate 95\% intervals across 5 replicate runs.
    }
    \label{fig:nyu-retention}
\end{figure}

We further evaluated the calibration of the uncertainties obtained from Bayesian inference
on PEFT subspaces by evaluating the test loss on quantiles of most certain pixels, 
as suggested in \cite{ciosek_conservative_2020}.
As an uncertainty metric, we consider the pixel-wise standard deviation. 
Results are depicted in \Cref{fig:nyu-retention} and \Cref{fig:kitti-retention}.
Note that for LoRA and CoLoRA, we only included the methods with smallest test loss on the
5\% quantile of most certain pixels.
Contrary to the previous section, we observe the worst performance for DeepEns,
where the test loss decreases instead of increasing on the most certain pixels.
However, similar to the results from the previous section, we observe the best performance
using DeepEns on the full parameter space.
For CkptEns and SWAG-D, inference on the full parameter space performs a bit worse than
BitFit and DiffFit, especially on the most certain pixels.
SWAG-LR overall achieves the smallest test loss on approximately 50\% of the most certain pixels
when using BitFit and DiffFit.
Quite interestingly, we observe BitFit to be slightly better calibrated than DiffFit, although
the latter uses more parameters.
On NYU (c.f.,~\Cref{fig:nyu-retention}), CoLoRA is outperformed by the other methods
when using SWAG,
however on KITTI (c.f.,~\Cref{fig:kitti-retention}), CoLoRA is often the only method
achieving test loss smaller than the baseline.

\begin{figure}
    \centering
    \includegraphics[width=\columnwidth]{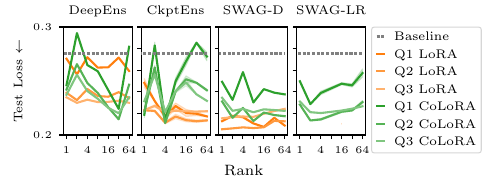}
    \caption{
        Test loss on 25\%, 50\%, and 75\% quantiles for LoRA and CoLoRA against
        the rank parameter, evaluated on the NYU data set.
        No clear trend suggesting the usage of higher ranks can be identified.
        Shaded areas indicate 95\% intervals across 5 replicate runs.
    }
    \label{fig:nyu-rank}
\end{figure}

Furthermore, we analyze the influence of the rank parameter on calibration
by considering the test loss at the 25\%, 50\%, and 75\% quantiles against
the rank parameters.
Results are depicted in \Cref{fig:nyu-rank} and \Cref{fig:kitti-rank}.
For both data sets, the results are fairly noisy and do not suggest
a clear trend favoring higher ranks and thus, more parameters.
Interestingly, rank $r=4$ seems to work particularly well for both 
data sets when using CkptEns.

\section{Conclusion}
\label{sec:conclusion}

We demonstrated the applicability of PEFT subspaces for Bayesian deep learning
on a modern Transformer-based computer vision architecture for monocular depth estimation.
We show that simple methods like checkpoint ensembles and SWAG are capable of
improving predictive performance and providing well-calibrated uncertainty estimates.
Moreover, we propose a novel approach for constructing LoRA-like subspaces in
convolutional layers, termed CoLoRA, and demonstrate that it performs competitively 
with the other PEFT methods.
We hope that CoLoRA can also serve to make existing, convolution-based architectures uncertainty-aware
in a parameter-efficient manner.

\section*{Acknowledgements}
\label{sec:acknowledgements}

RDP and AQ performed this work as part of the Helmholtz School
for Data Science in Life, Earth and Energy (HDS-LEE) and
received funding from the Helmholtz Association. 
RDP performed parts of this work as part of the HIDA Trainee Network program and 
received funding from the Helmholtz Information \& Data Science Academy (HIDA).
VF was supported by a Branco Weiss Fellowship.
The authors gratefully acknowledge computing time on the supercomputers
JURECA \cite{thornig2021jureca} and JUWELS \cite{kesselheim2021juwels} at Forschungszentrum J\"ulich.

\bibliography{literature}
\bibliographystyle{ieee_fullname}

\newpage
\appendix
\onecolumn

\section{Experiment Details}
\label{sec:details}

Fine-tuning is performed for 20 more epochs, starting from the checkpoints provided by Yang et al. \cite{yang_depth_2024}.
During fine-tuning take 100 equidistant checkpoints.
For both SWAG variants, we also draw 100 samples from the approximate posterior.
For LoRA and CoLoRA, which admit a rank parameter, we test ranks 1, 2, 4, 8, 16, 32 and 64.
For every combination of posterior approximation and PEFT method, 
we perform 5 replicate experiments using
different seeds.
For DeepEns, we use the last checkpoint from the five replicates to compile an ensemble.
For all experiments, we use Adam \cite{kingma_adam_2015} with a constant learning 
rate of 1e-7.
The batch size for all methods is set to 4.

\section{Results on KITTI.}
\label{sec:kitti}

\begin{figure}
    \begin{subfigure}{3.25in}
        \centering
        \includegraphics[width=3.25in]{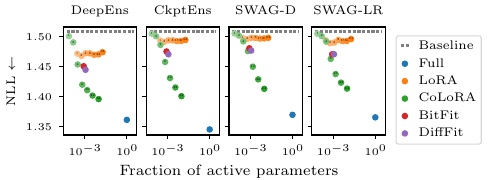}
        \caption{
            Negative log-likelihood and mean test loss for all combinations of inference
            and PEFT methods under consideration, evaluated on the KITTI data set. 
            Except SWAG-LR, all methods achieve improved NLL over the deterministic baseline.
            Error bars indicate 95\% intervals across 5 replicate runs.
        }
        \label{fig:kitti-error}
    \end{subfigure}~\hspace{.12in}~
    \begin{subfigure}{3.25in}
        \centering
        \includegraphics[width=3.25in]{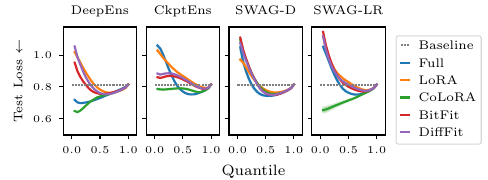}
        \caption{
            Test loss per quantile of most certain predictions evaluated on
            the KITTI data set.
            Except DeepEns, all methods achieve improved test loss on more certain
            pixels, suggesting good calibration.
            Uncertainty was estimated either using pixelwise standard deviation.
            For LoRA and CoLoRA, only the methods with lowest test loss on the
            5\% quantile are depicted.
            The prediction using the publicly available checkpoint was used as
            a baseline.
            Shaded areas indicate 95\% intervals across 5 replicate runs.
        }
        \label{fig:kitti-retention}
    \end{subfigure}
    \begin{center}
        \begin{subfigure}{3.25in}
            \vspace{.25in}
            \centering
            \includegraphics[width=3.25in]{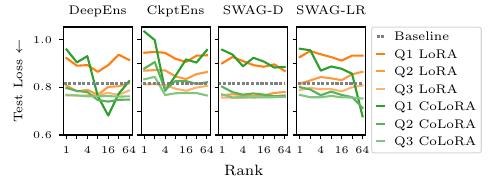}
            \caption{
                Test loss on 25\%, 50\%, and 75\% quantiles for LoRA and CoLoRA against
                the rank parameter, evaluated on the KITTI data set.
                No clear trend suggesting the usage of higher ranks can be identified.
                Shaded areas indicate 95\% intervals across 5 replicate runs.
            }
            \label{fig:kitti-rank}
        \end{subfigure}
    \end{center}
    \caption{Evaluations on KITTI data set.}
    \label{fig:kitti}
\end{figure}

We provide figures for the analyses from \Cref{sec:experiments} on the KITTI data set
in \Cref{fig:kitti}.

\section{Tucker Decomposition}
\label{sec:tucker}

For an $n$-tensor $A$ of size $h_1 \times \cdots \times h_n$,
the Tucker decomposition returns a core tensor $C \in \R^{r_1 \times \cdots \times r_n}$ and $n$ factor matrices
$U^{(1)}, \ldots, U^{(n)} \in \R^{h_i \times r_i}$, where $r_1,\ldots,r_n$ are the ranks along each
of the $n$ tensor dimensions.
From the Tucker decomposition, $A$ is recovered as
\eq{tucker-decomp}{
    a_{i_1,\ldots,i_n} = 
        \sum_{j_1=1}^{r_1} \cdots \sum_{j_n=1}^{r_n}
            c_{j_1,\ldots,j_n} \cdot u^{(1)}_{i_1,j_1} \cdots u^{(n)}_{i_n,j_n}.
}
A low-rank approximation of $A$ can be computed by choosing the ranks
$r_1, \ldots, r_n$ of the decomposition to be less than the full ranks.
Moreover, in the partial Tucker decomposition, we may choose to omit decomposition
of certain tensor dimensions of the incoming $A$ tensor, in which case the core
matrix $C$ simply takes full size $r_i=h_i$ along the respective tensor dimension $i$
and the corresponding factor matrix $U^{(i)} = I \in \R^{h_i \times h_i}$
becomes an identity matrix.
If the first $n$ tensor dimensions are chosen as low-rank, the corresponding decomposition is also
commonly called \emph{Tucker-$n$ decomposition}.

\end{document}